\pdfoutput=1

\documentclass[11pt]{article}

\usepackage{ACL2023}

\usepackage{times}
\usepackage{latexsym}

\usepackage[T1]{fontenc}

\usepackage[utf8]{inputenc}

\usepackage{microtype}

\usepackage{inconsolata}

\usepackage{amsmath,amsfonts,bm}

\def\eqref#1{equation~\ref{#1}}

\def\1{\bm{1}}

\def\mC{{\bm{C}}}

\def\mK{{\bm{K}}}
\def\mL{{\bm{L}}}
\def\mM{{\bm{M}}}

\def\mQ{{\bm{Q}}}
\def\mR{{\bm{R}}}
\def\mS{{\bm{S}}}

\def\mV{{\bm{V}}}
\def\mW{{\bm{W}}}

\def\mZ{{\bm{Z}}}

\DeclareMathAlphabet{\mathsfit}{\encodingdefault}{\sfdefault}{m}{sl}
\SetMathAlphabet{\mathsfit}{bold}{\encodingdefault}{\sfdefault}{bx}{n}

\newcommand{\R}{\mathbb{R}}

\usepackage{graphicx}
\usepackage{enumitem,kantlipsum}
\usepackage[hang,flushmargin]{footmisc}
\title{Implicit Memory Transformer for Computationally Efficient \\
Simultaneous Speech Translation}

\author{Matthew Raffel \and  Lizhong Chen \\ Oregon State University, USA \\ \texttt{\{raffelm, chenliz\}@oregonstate.edu}}

\begin{document}
\maketitle
\begin{abstract}
Simultaneous speech translation is an essential communication task difficult for humans whereby a translation is generated concurrently with oncoming speech inputs. For such a streaming task, transformers using block processing to break an input sequence into segments have achieved state-of-the-art performance at a reduced cost.  Current methods to allow information to propagate across segments, including left context and memory banks, have faltered as they are both insufficient representations and unnecessarily expensive to compute.  In this paper, we propose an \textit{Implicit Memory Transformer} that implicitly retains memory through a new left context method, removing the need to explicitly represent memory with memory banks.  We generate the left context from the attention output of the previous segment and include it in the keys and values of the current segment's attention calculation.  Experiments on the MuST-C dataset show that the Implicit Memory Transformer provides a substantial speedup on the encoder forward pass with nearly identical translation quality when compared with the state-of-the-art approach that employs both left context and memory banks.
\end{abstract}

\section{Introduction}
Simultaneous speech translation (SimulST) refers to the process of producing an output translation concurrently with an oncoming source speech input.  For humans, performing accurate SimulST is extremely difficult and becomes nearly impossible to perform over long periods of time.  Given the potential broad applications of SimulST in industry and government sectors, there is a strong need for machine learning models to perform the task to a level above the capabilities of humans.  

One branch of machine learning models that have been effective in SimulST is transformers \citep{vaswani2017attention} using block processing, a process that breaks an input sequence into segments which the encoder processes sequentially and individually \citep{dong2019self}.  As later segments may lose earlier information in a sentence (i.e., context fragmentation), techniques known as left context and memory banks have been introduced.  The concept of left context was idealized with the Transformer-XL \citep{dai2019transformer}, a model optimized for language modeling, which was later adapted for streaming automatic speech recognition (ASR).  The Transformer-XL generated left context by saving the previous segment to each encoder layer, so the subsequent segment could include it in the attention calculation at the same encoder layer.  Memory banks were later introduced in the self-attention calculation of the Augmented Memory Transformer \citep{wu2020streaming}, allowing it to outperform the Transformer-XL in streaming ASR and also be state-of-the-art in SimulST \citep{ma2021streaming}.  These memory banks were token summarizations of previous segments and helped retain explicit long-term dependencies. The Augmented Memory Transformer also included the left context alongside the center (main) segment tokens with an additional right context, all of which add computational cost.  We argue that the methods to generate and use left context and/or memory banks in both the Transformer-XL and Augmented Memory Transformer are naive, costing both models' performance at a given computational budget.

In this paper, we propose a computationally efficient architecture, the \textit{Implicit Memory Transformer}, that implicitly retains memory through a novel left context generation method, thereby removing the need for memory banks entirely.  Briefly, the proposed left context method for a given encoder layer leverages the previous segment's attention output in the attention calculation of the current segment.  Our method for calculating left context is broadly applicable to any transformer model that utilizes block processing. 
The proposed Implicit Memory Transformer is more computationally efficient than the Augmented Memory Transformer, reducing the cost of self-attention calculation, convolution layers, and feed-forward layers.  

We conduct our experiments on the English-German, English-French, and English-Spanish language pairs of the MuST-C dataset \citep{CATTONI2021101155} and demonstrate a significant speedup over the Augmented Memory Transformer for the forward pass of the encoder, with no reduction in the translation quality across all wait-$k$ values.

\section{Background and Related Works}
\textbf{Augmented Memory Transformer:} For SimulST, a transformer model waits for $k$ token chunks before beginning translation, a policy referred to as wait-$k$ \citep{ma2018stacl}.  One such transformer that uses this wait-$k$ policy is the Augmented Memory Transformer \citep{ma2021streaming}.  The Augmented Memory Transformer breaks an input sequence into segments $\mS_n^i \in \R^{s\times d}$, where $n$ denotes the segment position in the sequence and $i$ denotes the layer index in the Augmented Memory Transformer.  Each segment is composed of a left context $\mL_n^i \in \R^{l\times d}$ of size $l$, a center context $\mC_n^i \in \R^{c\times d}$ of size $c$, and a right context $\mR_n^i \in \R^{r\times d}$ of size $r$.  Each segment is of size $s=l+c+r$ and overlaps with the previous and subsequent segments with the left and right context.  Unlike the default transformer, the encoder of the Augmented Memory Transformer possesses two subsampling convolution layers to reduce the size of the segment inputs.   

In the self-attention calculation for the encoder, memory banks, $\mM_n^i \in \R^{N\times d}$, are added to the keys and values where $N$ denotes the maximum number of memory banks for a given layer. Each layer's memory banks summarize the previous segments and are theorized to allow the model to retain explicit long-term memory.  Each memory bank is created using the attention output of a summarization query, $\sigma_n^i\in \R^{1\times d}$, included in the attention calculation.  This summarization query is calculated by averaging the tokens in the current segment.  For any given layer, the queries, keys, and values can be represented by the following equations:
\begin{equation}
    \mQ_n^i = \mW_q^i[\mL_n^i, \mC_n^i, \mR_n^i, \sigma_n^i]
    \label{eq:query}
\end{equation}
\begin{equation}
    \mK_n^i = \mW_k^i[\mM_n^i,\mL_n^i,\mC_n^i, \mR_n^i]
    \label{eq:key}
\end{equation}
\begin{equation}
    \mV_n^i = \mW_v^i[\mM_n^i,\mL_n^i,\mC_n^i, \mR_n^i]
    \label{eq:value}
\end{equation}

In each of the equations, $\mW_q^i$, $\mW_k^i$, and $\mW_v^i$ are the query, key, and value projection matrices for layer $i$.  The $[.]$ operator concatenates $\mL_n^i$, $\mC_n^i$, $\mR_n^i$ with $\sigma_n^i$ or $\mM_n^i$.  After the encoder processes each individual segment, they are concatenated before being provided to a simultaneous decoder \citep{ma2020simulmt}.


\noindent \textbf{Average Lagging:} Average Lagging is one prominent metric to determine the efficacy of a SimulST model \citep{ma2018stacl}.  It denotes in milliseconds the lag between the output translation and the input source sequence \citep{ma2020simulmt}.  

\noindent \textbf{BLEU Score:} An equally important metric to evaluate a SimulST model is the BLEU score, which measures the translation similarity between the predicted output and the target output.  The BLEU score ranges from 0 to 1 and is often represented with percentages \citep{papineni-etal-2002-bleu}.
\section{Methods}
\subsection{Implicit Memory Transformer}
We propose an Implicit Memory Transformer that leverages a new left context generation method to retain an implicit memory of previous segments.  As such, we are able to remove the explicit memory provided by the memory banks that are expensive to compute in the Augmented Memory Transformer.  Our new implicit memory left context is unique at each layer of the encoder, whereby it is composed of a portion of the output from the self-attention calculation of the previous segment's center context.  

Specifically, suppose our implicit memory left context is denoted as $\mZ_n^i \in \R^{l\times d}$. Then, in the self-attention calculation of the Implicit Memory Transformer, the queries, keys, and values for each layer’s attention calculation can be calculated as follows:
\begin{equation}
    \mQ_n^i = \mW_q^i[\mC_n^i, \mR_n^i]
\end{equation}
\begin{equation}
    \mK_n^i = \mW_k^i[\mZ_n^i,\mC_n^i, \mR_n^i]
\end{equation}
\begin{equation}
    \mV_n^i = \mW_v^i[\mZ_n^i,\mC_n^i, \mR_n^i]
\end{equation} 

In comparison with the calculation of the queries, keys, and values of the current state-of-the-art Augmented Memory Transformer shown in Equation \ref{eq:query}, \ref{eq:key} and \ref{eq:value}, our Implicit Memory Transformer has three notable differences consisting of:
\begin{enumerate}[wide, labelindent=0pt, label=\textbf{\arabic*})]
    \item \textbf{Removed memory banks}: The memory bank terms in Equation \ref{eq:key} and \ref{eq:value} not only provide the model with explicit long-term memory but also introduce a recurrence mechanism to the transformer, which is a form of implicit memory. By removing memory banks and instead including the recurrence mechanism in the left context, we capture the benefits of this implicit memory without the additional cost to compute memory banks.
    \vspace{-0.2cm}
    \item \textbf{Attention-based left context}: In using the output from the attention calculation of the previous segment rather than the raw segment input as left context like the Transformer-XL, we are able to capture a \textit{learned} representation of the previous segment at a given layer.  This is similar to the Augmented Memory Transformer using the attention output associated with the summarization query as a memory bank.  However, since we do not compress the segment into a summarization query, we capture a more realistic representation.
    \vspace{-0.2cm}
    \item \textbf{Removed left context in the queries}: The Augmented Memory Transformer, includes the left context in each segment and, subsequently, the queries to allow it to generate a learned representation of the left context alongside the current segment.  However, since our Implicit Memory Transformer already has a saved learned representation of the left context for a given layer, it removes the need to include the left context in the segment.  
\end{enumerate}
  
From the above attributes, the self-attention calculation of the Implicit Memory Transformer becomes more efficient than that of the Augmented Memory Transformer, as memory banks are no longer included in the keys and values, and the left context and summarization query are removed from the queries.  
Furthermore, our Implicit Memory Transformer reduces the computation cost of the feed-forward neural network and the convolution subsampling layers, as they no longer need to process tokens contained in the left context.

\subsection{Complexity Analysis}
We will now perform complexity analysis for the self-attention and convolution subsampling layers in the Augmented Memory Transformer.  The complexity analysis of a convolution subsampling layer with a kernel size of one is identical to that for the linear transformations in the feed-forward network.  The self-attention layer has a complexity of $O(n^2 \cdot d)$ and the convolution layer has a complexity of $O(K \cdot n \cdot d^2)$ where $n$ is the input sequence length, $d$ is the hidden size, and $K$ is the kernel size \citep{vaswani2017attention}.  

The complexity of the self-attention layer of the old Augmented Memory Transformer would thus be $O((N+l+c+r)(l+c+r) \cdot d)$ and the complexity with the new method of calculating left context would be $O((c+r)(l+c+r) \cdot d)$.  Similarly the complexity of the convolution layers would change from $O(K \cdot (l+c+r) \cdot d^2)$ to $O(K \cdot (c+r) \cdot d^2)$.  Given the computational complexity decrease for all layers in the Augmented Memory Transformer with respect to the left context size and memory banks, it lends to the possibility of increasing the left context size for greater translation performance.

\section{Experimental Setup}
We conducted experiments on the English-German (en-de), English-French (en-fr), and English-Spanish (en-es) language pairs from the MuST-C dataset \citep{CATTONI2021101155}. The data preparation scripts for the MuST-C dataset are provided in Fairseq\footnote{\url{https://github.com/facebookresearch/fairseq}} \citep{ott2019fairseq,wang2020fairseqs2t}, whereby Kaldi is used to generate 80-dimensional log-mel filter bank features, and text is tokenized with a SentencePiece 10k unigram vocabulary.  The statistics of the training, development, and test set (tst-COMMON) for the English-German, English-French, and English-Spanish language pairs of the MuST-C dataset are provided in Table \ref{tab:MuST-C}.

\begin{table}[h]
\centering
\begin{tabular}{llll}
\hline
\textbf{Language Pair} & \textbf{Train} & \textbf{Dev} & \textbf{Test}\\
\hline
\verb|en-de| & \verb|250942| & \verb|1415| & \verb|2580|\\
\verb|en-fr| & \verb|275085| & \verb|1412| & \verb|2632| \\
\verb|en-es| & \verb|265625| & \verb|1316| & \verb|2502|\\
\hline
\end{tabular}
\caption{\label{tab:MuST-C}
The number of sentences in the train, development, and test (tst-COMMON) sets of the MuST-C dataset for the en-de, en-fr, en-es language pairs \citep{CATTONI2021101155}.
}
\end{table}

The architectures of the Augmented Memory Transformer and Implicit Memory Transformer trained were nearly identical, containing 33.1 M parameters \citep{ma2021streaming}.  Their encoders consisted of 12 layers beginning with two convolution layers with a combined subsampling factor of 4, followed by a feed-forward neural network.  Their decoders consisted of 6 layers.  Each of these layers has a hidden size of 256 with 4 attention heads. Relative positional encodings were applied to each self-attention layer with a clipping distance of 16 \citep{shaw2018self}.  Layer normalization was performed prior to each layer. Additionally, we trained each model with a wait-1, wait-3, wait-5, and wait-7 policy using a pre-decision ratio of 8 \citep{ma2020simulmt}.  We provide public access to a derivative of Fairseq containing our implementation for the Implicit Memory Transformer\footnote{\url{https://github.com/OSU-STARLAB/ImplicitMemory}}. 

All training was performed on a single V100-32GB.  The training process consisted of ASR pre-training followed by SimulST training. For SimulST training, the models were trained with label-smoothed cross-entropy loss, the Adam optimizer \citep{kingma2014adam}, and an inverse square root scheduler.  There was a warm-up period of 7500 updates where the learning rate of 0.0001, followed by a learning rate of 0.00035.  To regularize the model weights, we used a weight decay value of 0.0001, a dropout of 0.1, an activation dropout of 0.2, and an attention dropout of 0.2.  All models were trained with early stopping using a patience of 10.  After the training was complete, the final ten checkpoints were averaged. 

The translation quality and latency were determined by detokenized BLEU with SacreBLEU \citep{post2018call}, and Average Lagging \citep{ma2020simulmt}, respectively.  
Both of these metrics were obtained using the SimulEval toolkit\footnote{\url{https://github.com/facebookresearch/SimulEval}}, which simulates SimulST \citep{ma-etal-2020-simuleval}.

\section{Results}
\subsection{Performance Evaluation}
\begin{figure}[ht]
\begin{center}
\centerline{\includegraphics[width=\columnwidth]{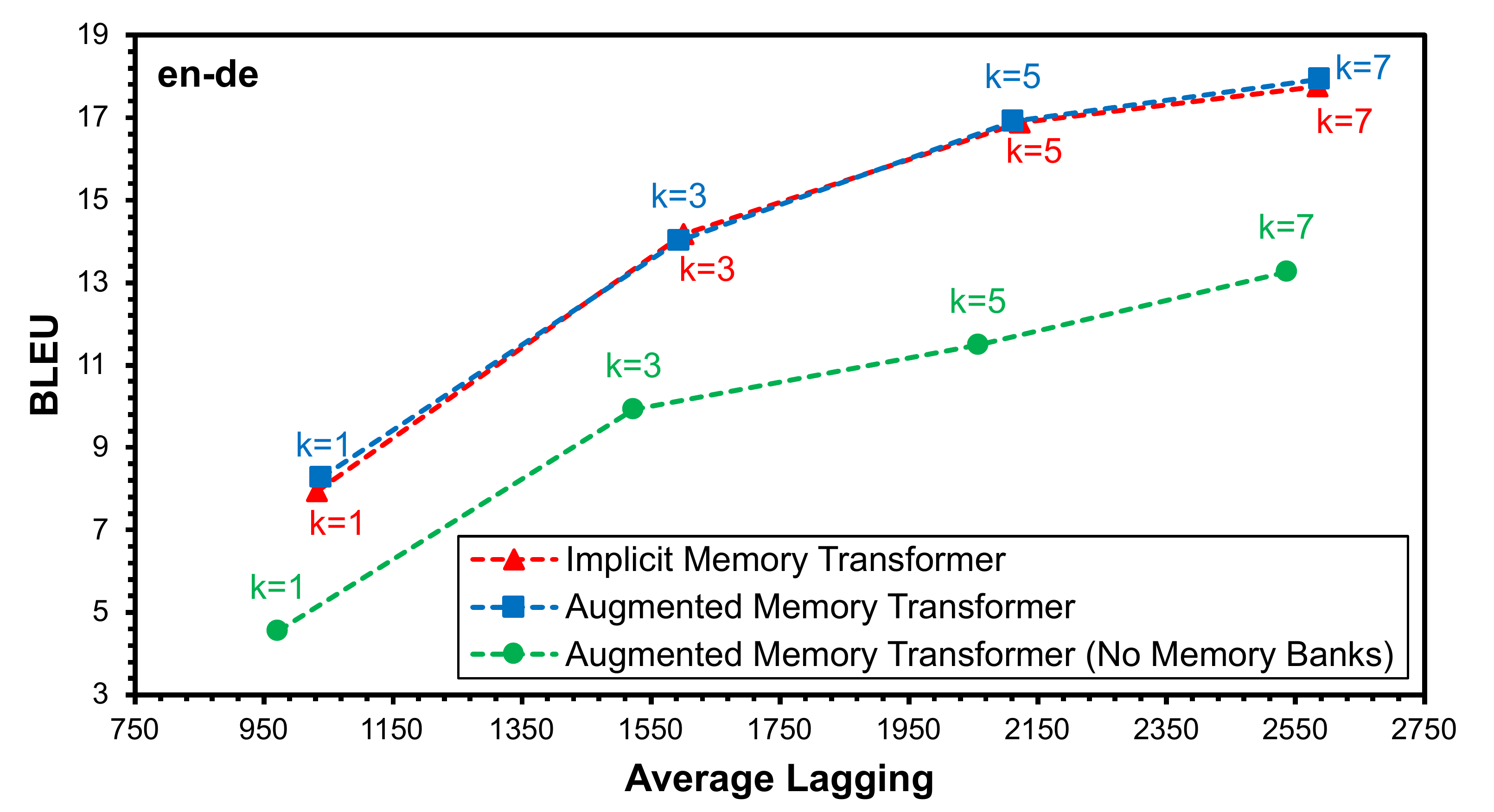}}
\caption{A comparison between the Implicit Memory Transformer and the baseline Augmented Memory Transformers on the \textbf{en-de} language pair.}
\label{fig:en-de}
\end{center}
\vskip -0.25in
\end{figure}

We demonstrate the efficacy of our Implicit Memory Transformer on the English-German language pair for a single run in Figure \ref{fig:en-de} in terms of average lagging and BLEU score.  Figure \ref{fig:en-de} compares our Implicit Memory Transformer against two Augmented Memory Transformers differing by the inclusion or exclusion of memory banks. Each of the compared models consists of a left context of 32 tokens, a right context of 32 tokens, and a center context of 64 tokens.  The Augmented Memory Transformer using memory banks have a total of three banks, whereas our Implicit Memory Transformer and the Augmented Memory Transformer without memory banks have zero. 

From viewing Figure \ref{fig:en-de}, our Implicit Memory Transformer achieves almost identical performance in terms of BLEU score to the Augmented Memory Transformer using memory banks for SimulST between English and German without affecting the Average Lagging.  In contrast, simply removing memory banks in the Augmented Memory Transformer results in an average 4.48 BLEU decrease versus its memory bank counterpart across all wait-$k$ values.  This confirms the effectiveness of the attention-generated left context of the proposed Implicit Memory Transformer for the English-German language pair.

\begin{figure}[ht]
\begin{center}
\centerline{\includegraphics[width=\columnwidth]{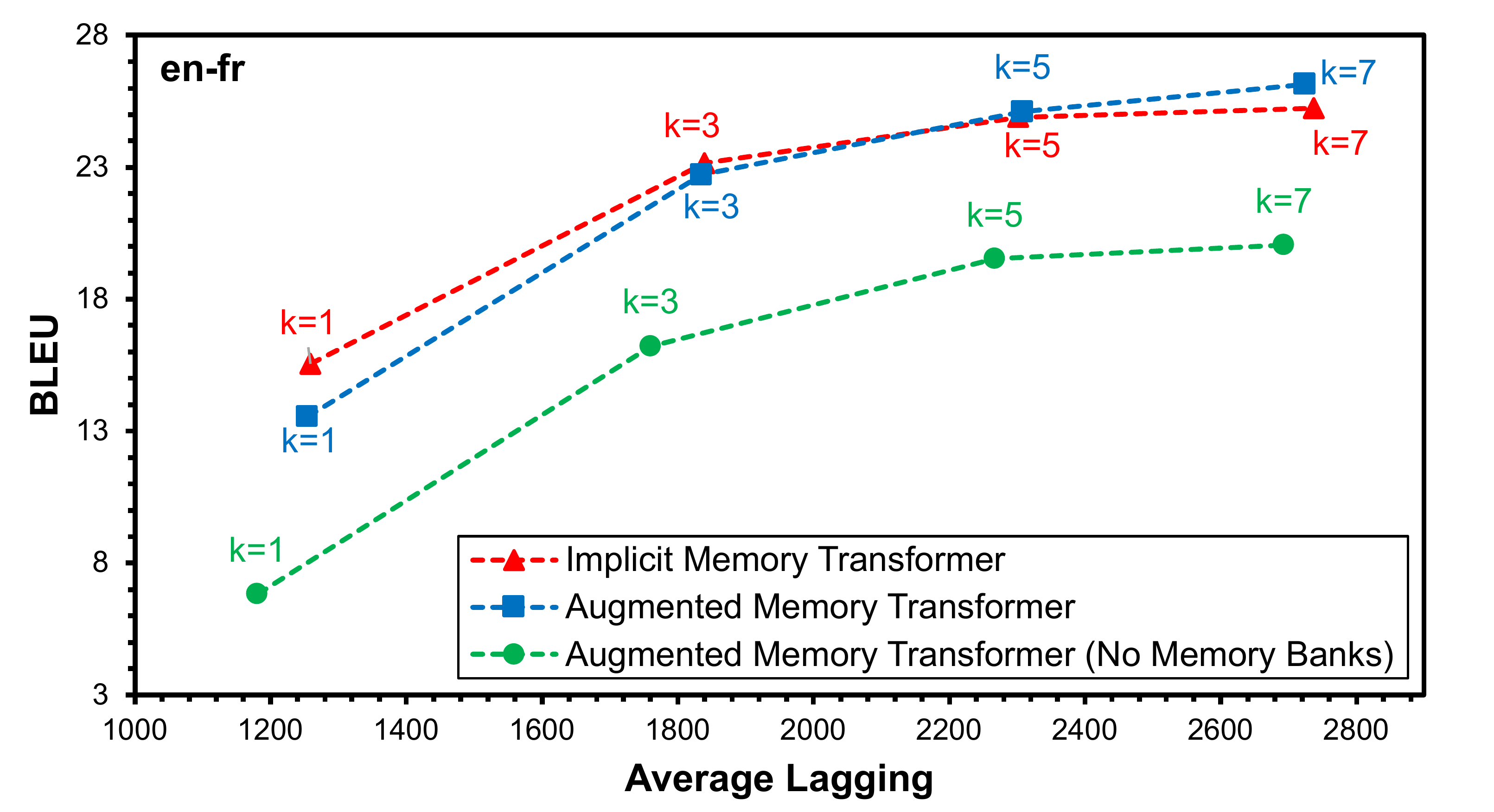}}
\caption{A comparison between the Implicit Memory Transformer and the baseline Augmented Memory Transformers on the \textbf{en-fr} language pair.}
\label{fig:en-fr}
\end{center}
\vskip -0.25in
\end{figure}

\begin{figure}[ht]
\begin{center}
\centerline{\includegraphics[width=\columnwidth]{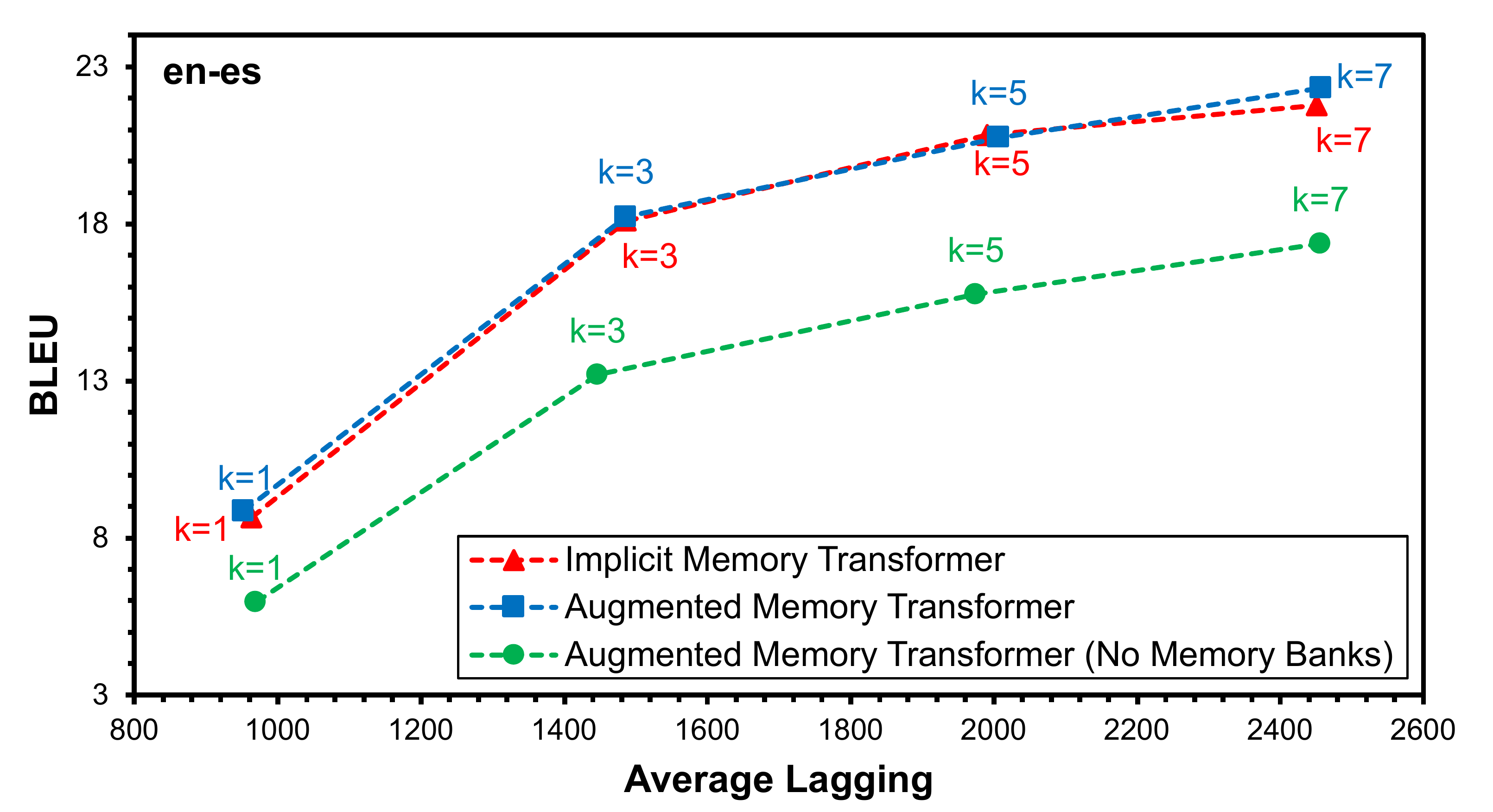}}
\caption{A comparison between the Implicit Memory Transformer and the baseline Augmented Memory Transformers on the \textbf{en-es} language pair.}
\label{fig:en-es}
\end{center}
\vskip -0.25in
\end{figure}

We see similar results with the English-French and English-Spanish language pairs provided in Figure \ref{fig:en-fr} and Figure \ref{fig:en-es}, respectively.

In both cases, the Implicit Memory Transformer performs nearly identically to the Augmented Memory Transformer using memory banks by not negatively impacting either the BLEU score or Average Lagging.  Additionally, as with the results in Figure \ref{fig:en-de}, the Augmented Memory Transformer sees an average decrease of 6.23 BLEU and 4.47 BLEU across all wait-$k$ values when memory banks are removed for the English-French and English-Spanish language pairs respectively.  Once again substantiating the efficacy of our attention-generated left context in the Implicit Memory Transformer, which does not see a performance decrease without memory banks. 

\subsection{Evaluation Speedup}
\begin{figure}[ht]
\begin{center}
\centerline{\includegraphics[width=\columnwidth]{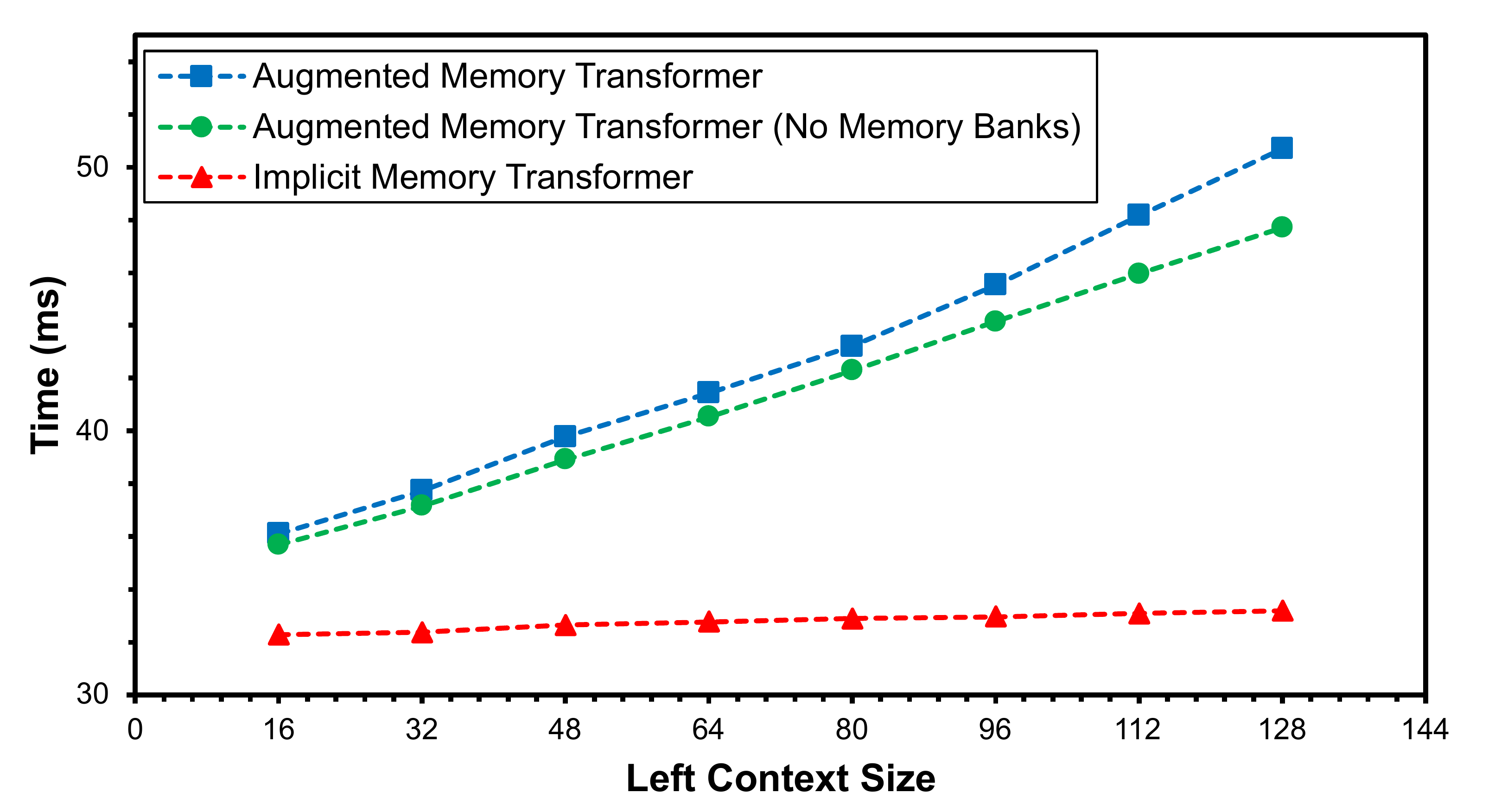}}
\caption{A comparison between the forward pass time (ms) of the Implicit Memory Transformer and two Augmented Memory Transformer variations with respect to the left context size.}
\label{fig:speed}
\end{center}
\vskip -0.25in
\end{figure}

We provide a demonstration of how the left context size affects the forward pass time of a segment through the encoder of an Augmented Memory Transformer with three memory banks, an Augmented Memory Transformer without memory banks, and the Implicit Memory Transformer in Figure \ref{fig:speed}.  The left context size is scaled with tokens, and the duration of the forward pass of a segment through the encoder is scaled in milliseconds.  Each model compared uses a right context of 32 tokens and a center context of 64 tokens for each tested left context size.  Each measurement point in Figure \ref{fig:speed} is made by averaging the duration of ten forward passes through the encoder using two 14-core 2.20 GHz Intel Xeon Gold 5120 with 19712 KB cache. 

In Figure \ref{fig:speed}, the forward pass time of the Implicit Memory Transformer remains flat with respect to the left context size, whereas the two Augmented Memory Transformer models exhibit a nonlinear curved relationship.  
The separation of two Augmented Memory Transformer curves becomes more apparent for larger left context sizes, indicating the cost of memory banks becomes more apparent as the segment size increases.  
Figure \ref{fig:speed} also shows that dropping the left context from the query in the proposed Implicit Memory Transformer achieves considerable additional computation reduction beyond removing memory banks.

\section{Conclusion}
Achieving computationally efficient simultaneous speech translation (SimulST) is critical to its deployment in practical real-time applications. However, even with the state-of-the-art SimulST approach of the Augmented Memory Transformer, its method of generating left context is computationally costly and ineffective, requiring the usage of memory banks to compensate for its shortcomings.  As such, we propose an Implicit Memory Transformer that utilizes an attention-based left context to provide the model with implicit memory.  
We found that the Implicit Memory Transformer was able to achieve nearly identical performance to the Augmented Memory Transformer at a significantly reduced computational cost.

\section*{Limitations}
Our work is limited as it has not explored the effectiveness of our Implicit Memory Transformer in other tasks outside of SimulST, such as ASR.  We have also not explored the impact of our implicit memory left context on alternative block-processing-based transformer models.  Furthermore, extensive ablation studies could help showcase the potential of the Implicit Memory Transformer. 

\section*{Acknowledgements}
This research was supported, in part, by the National Science Foundation grants 2223483 and 2223484.

\bibliography{anthology,custom}
\bibliographystyle{acl_natbib}


\end{document}